\documentclass[conference]{IEEEtran}
\IEEEoverridecommandlockouts
\usepackage{cite}
\usepackage{amsmath,amssymb,amsfonts}
\usepackage{algorithmic}
\usepackage{graphicx}
\usepackage{textcomp}
\usepackage[rgb]{xcolor}
\usepackage{pgfplots}
\usepackage{pgfplotstable}
\usepackage{caption}
\usepackage{subcaption}
\usepackage{tcolorbox}
\usepackage{mathtools}
\usepackage{amsthm}
\usepackage{booktabs}
\usepackage{makecell}

\newcommand{\wname}{\textsc{Script Writer}}
\newcommand{\vname}{\textsc{Video Generator}}

\theoremstyle{plain}

\theoremstyle{definition}

\theoremstyle{remark}

\def\BibTeX{{\rm B\kern-.05em{\sc i\kern-.025em b}\kern-.08em
    T\kern-.1667em\lower.7ex\hbox{E}\kern-.125emX}}

\let\oldtwocolumn\twocolumn
\renewcommand\twocolumn[1][]{%
    \oldtwocolumn[{#1}{
    \begin{center}
           \includegraphics[width=\textwidth]{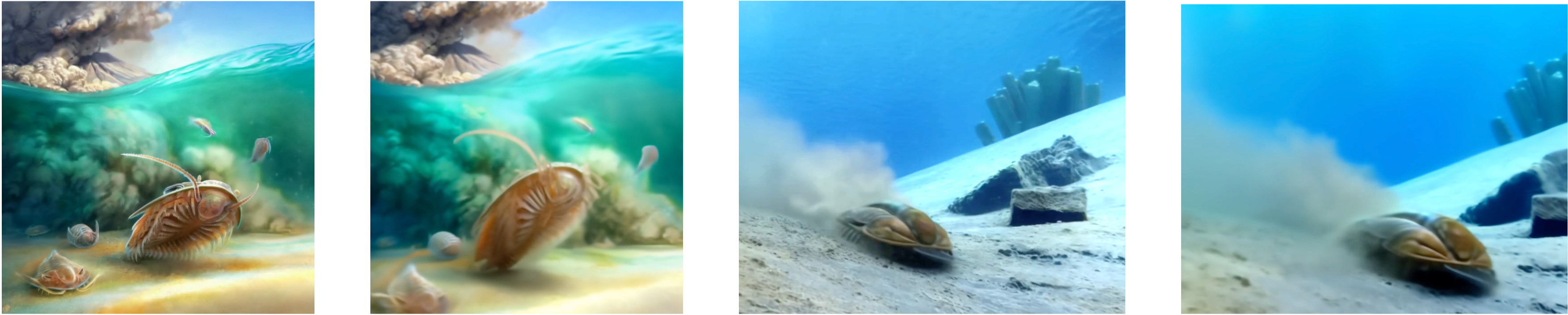}
           \captionof{figure}{We design and train the first text-to-video framework that automatically learn to refine the prompts to generate visually realistic trilobites adhering closely to pronounced and authentic trilobite characteristics in more fluid and lifelike videos. The first prompting image is courtesy of \cite{el2024rapid}.}
           \label{fig:fig1}
        \end{center}
    }]
}
\begin{document}

\title{Animating the Past: \\Reconstruct Trilobite via Video Generation
}

\author{\IEEEauthorblockN{Xiaoran Wu\IEEEauthorrefmark{1}}
\IEEEauthorblockA{
\textit{AI Lab} \\
\textit{Yishi Inc.}\\
Hangzhou, China \\
wuxr18@tsinghua.org.cn}
\and
\IEEEauthorblockN{Zien Huang\IEEEauthorrefmark{1}}
\IEEEauthorblockA{\textit{The International Department} \\
\textit{Experimental High School}\\
\textit{Attached to Beijing Normal University}\\
keane.huangzien@gamil.com}
\and
\IEEEauthorblockN{Chonghan Yu}
\IEEEauthorblockA{\textit{School of Ocean Sciences} \\
\textit{China University of Geoscience}\\
Beijing, China \\
yu\_chonghan@foxmail.com}
}

\maketitle
\begingroup
\renewcommand\thefootnote{\fnsymbol{footnote}}
\footnotetext[1]{These authors contributed equally to this work.}
\endgroup

\begin{abstract}
Paleontology, the study of past life, fundamentally relies on fossils to reconstruct ancient ecosystems and understand evolutionary dynamics. Trilobites, as an important group of extinct marine arthropods, offer valuable insights into Paleozoic environments through their well-preserved fossil records. Reconstructing trilobite behaviour from static fossils will set new standards for dynamic reconstructions in scientific research and education. Despite the potential, current computational methods for this purpose like text-to-video (T2V) face significant challenges, such as maintaining visual realism and consistency, which hinder their application in science contexts. To overcome these obstacles, we introduce an automatic T2V prompt learning method. Within this framework, prompts for a fine-tuned video generation model are generated by a large language model, which is trained using rewards that quantify the visual realism and smoothness of the generated video. The fine-tuning of the video generation model, along with the reward calculations make use of a collected dataset of 9,088 \emph{Eoredlichia intermedia} fossil images, which provides a common representative of visual details of all class of trilobites. Qualitative and quantitative experiments show that our method can generate trilobite videos with significantly higher visual realism compared to powerful baselines, promising to boost both scientific understanding and public engagement.
\end{abstract}

\IEEEpeerreviewmaketitle


\section{Introduction}

Paleontology, the study of prehistoric life, relies heavily on the fossil record to reconstruct past ecosystems, understand evolutionary processes, and decipher extinct organisms' biology~\cite{fortey2014palaeoecology,trilobite2019trilobites}. As an extinct group of marine arthropods, trilobites are among the most iconic and well-studied fossils~\cite{bergstrom1973organization,hughes2007evolution,fortey2014palaeoecology}, providing critical insights into Paleozoic ecosystems. Reconstructing the behavior and locomotion of trilobites is of great research and educational interests~\cite{el2024rapid}. Such dynamic reconstructions help in formulating hypotheses about trilobites' living environments and the functional morphology and ecological roles of these ancient creatures~\cite{levi1995trilobites,bergstrom1973organization,hughes2007evolution,ou2009juvenile}. Furthermore, from the educational aspect, reconstruction provides a tangible visualization of trilobite appearance and behavior, thus bridging the gap between abstract scientific knowledge and public understanding~\cite{hopkins2017development}.

Despite the abundance in the trilobite fossil record, reconstructing their behavior and movement remains a challenge, primarily due to the limitations inherent in fossil remains, such as their static nature. Fortunately, recent advancements in generative artificial intelligence (AI) and computational techniques provide new opportunities to address these challenges~\cite{hunt2014artificial, lecun2015deep,singer2022make,khachatryan2023text2video,wu2023tune}. Integrating AI into paleontological research not only showcases the potential of extending machine learning into a field of natural research that AI has not studied extensively before~\cite{sohrabi2021artificial} but also hopefully can enhance our understanding of the trilobite ethology and shed new light on its study.

Among generative AI techniques, video generation~\cite{el2024rapid,li2018video,hong2022cogvideo} techniques in particularly suitable for simulating trilobite movement in a dynamic, visually engaging manner. However, the current methods of video generation encounter several challenges that hinder their application to paleontological reconstructions. Primarily, as demonstrated in our qualitative studies, existing methods struggle with maintaining the realism of the depicted trilobites, with the creatures appearing unrealistic or oddly shaped~\cite{trilobite2019trilobites}. This lack of realism significantly detract from the viewer's engagement and reduce the educational and research value of the visualizations. Moreover, the consistency of generated videos often falls short, with noticeable discrepancies between consecutive frames~\cite{ren2024consisti2v, henschel2024streamingt2v}. Such inconsistencies are particularly problematic in longer sequences, leading to choppy transitions that disrupt the fluid simulation of trilobite movement. 

To tackle these issues, we propose a novel approach that embeds the evaluation of trilobite realism and video smoothness directly into the video generation workflow. Our solution leverages diffusion models~\cite{kingma2021variational,yang2023diffusion,ho2022video, croitoru2023diffusion}, which have demonstrated impressive capabilities in producing realistic images and videos from textual descriptions. We employ these models to create a series of animated segments that capture various aspects of trilobite movement, guided by descriptive prompts generated by a large language model (LLM)~\cite{achiam2023gpt, jiang2024mixtral, touvron2023llama}. The cornerstone of our method involves assessing the smoothness of transitions and the accuracy of the trilobite appearance in these animations, compared against a curated collection of trilobite fossil images. This assessment acts as a feedback mechanism to fine-tune the LLM that generates prompts for the text-to-animation model~\cite{bouali2023review, guo2023animatediff, hu2024animate}, enhancing the fidelity of animations. The objective is dual: to produce animations that not only accurately depict trilobite appearance and movement but also ensure seamless transitions, adding a layer of complexity to the model's training but crucial for high-quality video output.

In summary, our methodology encompasses several stages: initially, we generate basic animated segments from a set list of LLM-generated prompts. These segments are then pieced together, and the composite video is evaluated for the quality of its transitions and the realism of its content. The evaluation results are used as reward signals to update the LLM with preference optimization~\cite{rafailov2024direct,ethayarajh2024ktomodelalignmentprospect} to refine the animations. This cycle of generation, evaluation, and enhancement is repeated until the video meets our criteria for smoothness and realism.

We comprehensively evaluate our method both qualitatively and quantitatively against state-of-art text-to-animate and text-to-video academic research~\cite{guo2023animatediff} and commercial tools~\cite{pika,gen2}. The results show clear advances in paleontological visualization in terms of content realism and video continuity. Furthermore, we provide ablation studies to show the contribution of each component in our learning framework. We hope that this pioneering integration of technology and paleontology makes significant contributions to the field of synthetic media generation and opens new pathways for visualizing and understanding prehistoric entities and exploring ancient life.

\section{Related Work}
Our method of training the Large Language Model~(LLM) that generates prompts is related to Reinforcement Learning from Human Feedback (RLHF), an important technique to ensure that LLM outputs align with human preferences~\cite{ouyang2022training,bai2022training,wang2024rlhf}. 
Here in our work, the counterpart of human preference is defined by metrics regarding content realism and video continuity. Typically, RLHF initially learns a reward model~(RM)~\cite{gao2023scaling} from human preferences and then optimizes the supervised fine-tuned LLM model with reinforcement learning algorithms, e.g., PPO~\cite{schulman2017proximal}, to maximize the cumulative rewards from the RM. However, training the reward model is time-consuming and computation-intensive~\cite{gao2023scaling}. Direct Preference Optimaztion~(DPO)~\cite{rafailov2024direct} avoids training the reward model by directly aligning LLMs to best satisfy human preferences using a simple classification objective. The recently proposed KTO~\cite{ethayarajh2024ktomodelalignmentprospect} extends DPO by maximizing utility functions derived from prospect theory~\cite{tversky1992advances} for accurate human utility modelling. In this work, to achieve better stability and robustness, we utilize the calculated realism and continuity rewards to order different LLM outputs (prompts to the animation generation model) and use KTO for preference optimization.

Video generation methods like Tune-a-Video~\cite{wu2023tune} extend text-to-image (T2I) models to generate multiple images simultaneously by incorporating a tailored spatio-temporal attention mechanism and an efficient one-shot tuning strategy to learn continuous motion among generated images.
Text2Video-Zero~\cite{khachatryan2023text2video} proposes a cost-effective approach that requires no training or optimization by leveraging the capabilities of existing T2I synthesis methods, adapting them for the video generation domain. CogVideo~\cite{hong2022cogvideo} proposes a multi-frame-rate hierarchical training strategy to better align text and video clips on large-scale text-video datasets. Furthermore, commercial video generation tools are setting significant benchmarks. In this paper, we empirically compare our method against Pika~\cite{pika} and Gen3~\cite{gen2} for evaluation. Text-to-Animation~(T2A) is another approach in video generation that extends pre-trained T2I models by incorporating temporal structures~\cite{bouali2023review,guo2023animatediff}. AnimateDiff~\cite{guo2023animatediff} introduces a plug-and-play motion module that enables the training of T2A models without the need for model-specific tuning. In this paper, we employ the T2A method to generate trilobite animations from user prompts, but the focus is on how to enhance the temporal coherence and content realism.

We now introduce the preliminaries of the RLHF and T2A techniques based on which we develop our method.

\section{Preliminaries}


\textbf{RLHF}.\ \ The training of modern Large Language Models (LLMs) involves three phases as outlined in~\cite{jiang2023mistral, achiam2023gpt,ouyang2022training,wang2024q}. \emph{(1) Pretraining:} This phase involves training an initial model $\pi_0$ on a large text corpus to optimize the prediction of the next token based on the preceding text~\cite{bi2024deepseek, wei2023skywork, wei2024skywork}. \emph{(2) Supervised Fine-tuning (SFT):} The model is further trained on task-specific data that generally includes targeted instructions and expected responses, to refine its utility for practical applications~\cite{lu2023instag}. This fine-tuned model is noted as $\pi_{\text{ref}}$. \emph{(3) RLHF:} This step uses a preference dataset $\mathcal{D}$ containing tuples $(x, y_w, y_l)$ where $x$ is the input and $y_w$, $y_l$ are the preferred and less preferred outputs, respectively~\cite{bai2022training, rafailov2024direct}. A Bradley-Terry model~\cite{bradley1952rank} is used here to calculate preferences:
\[ p^*(y_w > y_l | x) = \sigma(r^*(x, y_w) - r^*(x, y_l)), \]
where $\sigma$ is the logistic function.
A reward model $r_{\phi}$ is trained by minimizing the negative log-likelihood of the preference data in the set $\mathcal{D}$~\cite{ouyang2022training}:
\[ \mathcal{L}_R(r_{\phi}) = \mathbb{E}_{(x,y_w,y_l) \sim \mathcal{D}}[- \log \sigma(r_{\phi}(x, y_w) - r_{\phi}(x, y_l))]. \]
To balance reward maximization with linguistic correctness, a KL divergence penalty is applied, preventing the model from deviating excessively from the reference model $\pi_{\text{ref}}$:
\[\max_{\pi_\theta} \mathbb{E}_{x\sim\mathcal{D},y \sim \pi_\theta(\cdot|x)} \left[r_{\phi}(x, y)\right] - \beta \mathbb{D}_\text{KL}\left[\pi_\theta(\cdot|x) || \pi_{\text{ref}}(\cdot|x)\right]. \]
The objective, non-differentiable in nature, requires an RL approach like PPO~\cite{schulman2017proximal} for optimization.

The computational demands and instability of training the reward model $r_\phi$ have led to the development of Direct Preference Optimization (DPO)~\cite{rafailov2024direct}, providing a stable alternative that trains directly on preference pairs and with similar optimal policy convergence performance:
\begin{align}
    &\mathcal{L}_{\text{DPO}}(\pi_\theta, \pi_{\text{ref}}) = \mathbb{E}_{(x,y_w,y_l) \sim \mathcal{D}}\\
    &\left[- \log \sigma\left(\beta \log \frac{\pi_\theta(y_w|x)}{\pi_{\text{ref}}(y_w|x)} - \beta \log \frac{\pi_\theta(y_l|x)}{\pi_{\text{ref}}(y_l|x)}\right)\right].\nonumber
\end{align}


\textbf{T2A}.\ \ One approach for short video generation is to animate a text-to-image (T2I)~\cite{li2019controllable} model by integrating temporal dynamics. Following the methodologies outlined in related literature~\cite{guo2023animatediff,hong2022cogvideo}, a batch of video data is represented as 5-dimensional tensors $x \in \mathbb{R}^{b \times c \times f \times h \times w}$. Here, $b$ denotes the batch axis, $f$ represents the frame-time axis, and $c$, $h$, and $w$ are the channels, height, and width of each video frame, respectively. The text-to-animation process begins by encoding each frame of a video data batch $x_{1:f
} \in \mathbb{R}^{b \times c \times f \times h \times w}$ into latent representations $z_{1:f,0}$ using a pre-trained auto-encoder. These representations are subsequently perturbed by noise as per the forward diffusion schedule~\cite{yang2023diffusion,song2021maximum}:
\begin{equation}
    z_{1:f,t} = \sqrt{\bar{\alpha}_t} z_{1:f,0} + \sqrt{1 - \bar{\alpha}_t} \epsilon, \quad \epsilon \sim \mathcal{N}(0, I), \quad t = 1, \ldots, T.
    \label{eq:3}
\end{equation}

In this inflated model, the noisy latent representations, along with corresponding text prompts, serve as inputs for predicting the noise added during the diffusion process. The training objective for T2As can be formulated as:
\begin{equation}
    \mathcal{L} = \mathbb{E}_{\mathcal{E}(x_{1:f}), y, \epsilon \sim \mathcal{N}(0, I), t} \left[ \left\| \epsilon - \epsilon_\theta (z_{1:f,t}, t, \tau_\theta(y)) \right\|^2_2 \right].
    \label{eq:4}
\end{equation}

By inflating the model with the additional temporal axis~\cite{shen2023mostgan}, this formulation emphasizes the critical role of temporal coherence and dynamic content adaptation in generating animations from textual descriptions.

However, this approach may yield animations that are not only brief in duration but also suffer from less smooth transitions between frames and inconsistencies in object appearances across different frames. These issues primarily arise from the model's limitations in maintaining consistent motion patterns and visual quality throughout the sequence~\cite{ren2024consisti2v}. This challenge is particularly pronounced when the model attempts to interpolate complex dynamics, a task that demands high fidelity in temporal and spatial representations. The difficulty lies in the model’s capacity to accurately generate and link successive frames where each must evolve naturally from its predecessor while adhering to the dynamics specified by the textual description.

\section{Method}

In this section, we describe our method that addresses the challenge of maintaining motion smoothness and visual realism throughout the video sequences.

The proposed framework synergizes the power of a large language model (\wname) that generates prompts and a fine-tuned text-to-animation model (\vname). Our main technical novelty lies in the design of the optimization algorithm of \wname. In effect, we design a contextual bandit learning task for the \wname. The concept of contextual bandit is popular in the cutting-edge LLM research, such as direct preference optimization (DPO~\cite{rafailov2024direct}).

For fine-tuning the \vname~and training the \wname, we collect a set $\mathcal{R}$ of 9088 \emph{Eoredlichia intermedia} fossil images, which include a large number of specimens covering different stages of individual development of \emph{Eoredlichia intermedia} trilobites. The images are used to provide a common representative of the visual traits of all class of trilobites in order to enhance the visual details of the generated content. These real trilobite fossil images do not mean that the videos produced in this study can fully reproduce the real trilobite structure, but the use of these real fossil details can greatly supplement the scarcity and errors of trilobite images in the web footage, enhancing the trilobite structure and details in the videos.

\subsection{Prompt and Video Initialization}
We now describe the details of our method. As the first step, the \wname~$\pi(\theta^0)$, where $\theta^0$ is the initial parameters, is asked to generate an initial prompt $y^0$ for the text-to-animation model with the format
\begin{align}
    y^0=(t_1: y^0_1; t_2: y^0_2;, \cdots, t_N, y^0_N),
\end{align}
where $y^0_n, n\in[N]$ is a textual description of the appearance and expected movement of a trilobite in the animation clip $n$, and $t_n, n\in[N]$ is the frame index from which the animation clip $n$ will start in the final video.

This initial prompt $y$ directs the text-to-animation diffusion model \vname~to generate $N$ initial animation clips $(c^0_1(y^0), \cdots, c^0_N(y^0))$ that are concatenated sequentially to get an initial video $z^0_{1:f}(y^0)$. Before generating this initial video, the \vname~has been fine-tuned on the collected fossil image dateset $\mathcal{R}$ to enhance the model's ability to generate the detailed textures and structures observed in fossil images.


\begin{figure*}
    \centering
    \includegraphics[width=\linewidth]{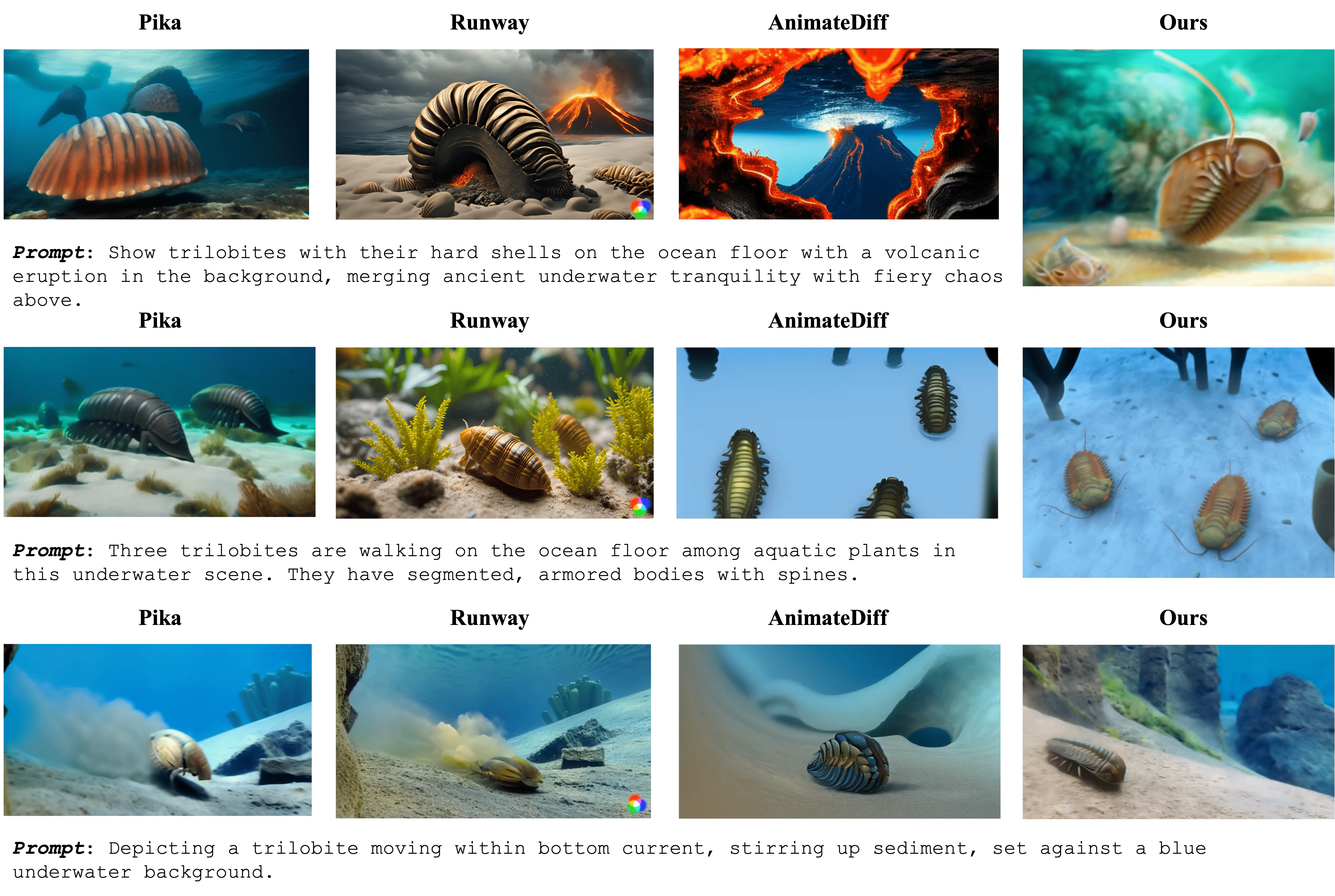}
    \caption{Qualitative comparison of the generated videos from four different models: Pika~\cite{pika}, Runway~\cite{gen2}, AnimateDiff~\cite{guo2023animatediff}, and ours. Our model significantly outperforms the others in generating trilobites with highly detailed morphological accuracy, realistic texturing, and appropriate environmental interactions. The prompting images in the first and second rows are courtesy of \cite{el2024rapid} and \cite{exampleImage3}, respectively.}
    \label{fig:real_compare_1}
\end{figure*}

\subsection{Reward Design for the \wname}
The second step is to design reward signals to train the \wname~with the hope that it can refine the initial prompt $y^0$ so that the resulting video has a better quality in terms of transition smoothness and visual realism. 

To be specific, the reward for prompt $y^0$ is designed to be a summation of two components: $r(y^0) = r_s(y^0)+r_a(y^0)$, where $r_s$ measures the smoothness of frame transition and $r_a$ measures the visual realism of the trilobites in the generated video.
 
\textbf{Smoothness of Frame Transition}.\ \ To assess the smoothness of a video, we compute the Fréchet Inception Distance (FID)~\cite{kynkaanniemi2022role} between adjacent frames. Consider two frames \(x_{t} \in \mathbb{R}^{ c \times h \times w}\) and \(x_{t+1} \in \mathbb{R}^{c \times h \times w}\), where \(c\), \(h\), and \(w\) represent the channel, height, and width of the frame, respectively. We first use a pre-trained InceptionV3 network~\cite{szegedy2016rethinking} to extract the image features (pool3 layer) \(z_{t}\) and \(z_{t+1}\) given frames \(x_{t}\) and \(x_{t+1}\), and then
compute the FID score for consecutive frames by:
    \begin{equation}
    \textsc{FID}_{t} = || z_{t} - z_{t+1} ||^2.
        \label{eq:5}
    \end{equation}

After obtaining the FID scores for all consecutive frames, we get the transition smoothness reward $r_s(y^0) = -\sum_{t=1}^f \textsc{FID}_t$. For fine-grained control, the reward can be calculated for each clip separately:
\begin{align}
    r_s(y^0_n) = -\sum_{t=t_n}^{t_{n+1}} \textsc{FID}_t.
\end{align}

\begin{figure*}
    \centering
    \includegraphics[width=\linewidth]{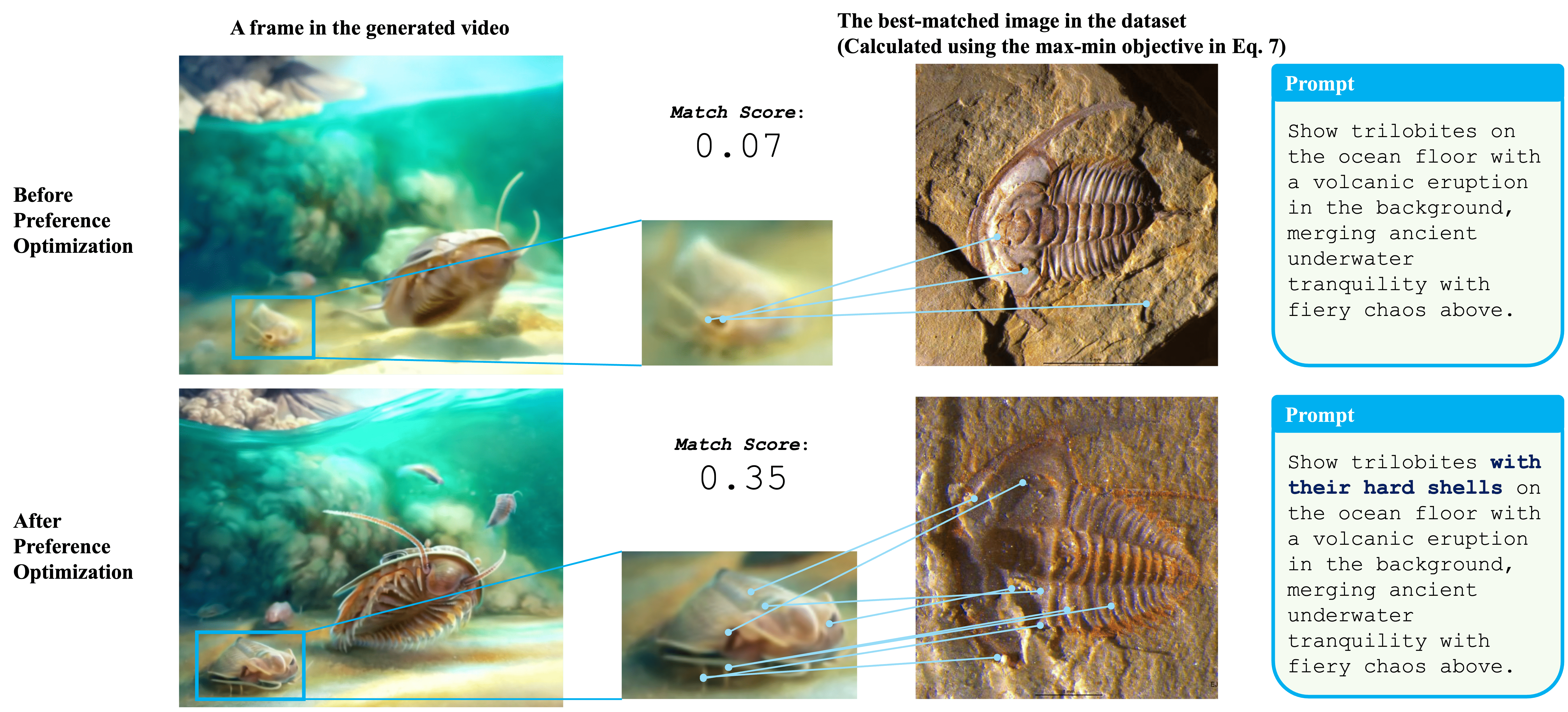}
    \includegraphics[width=\linewidth]{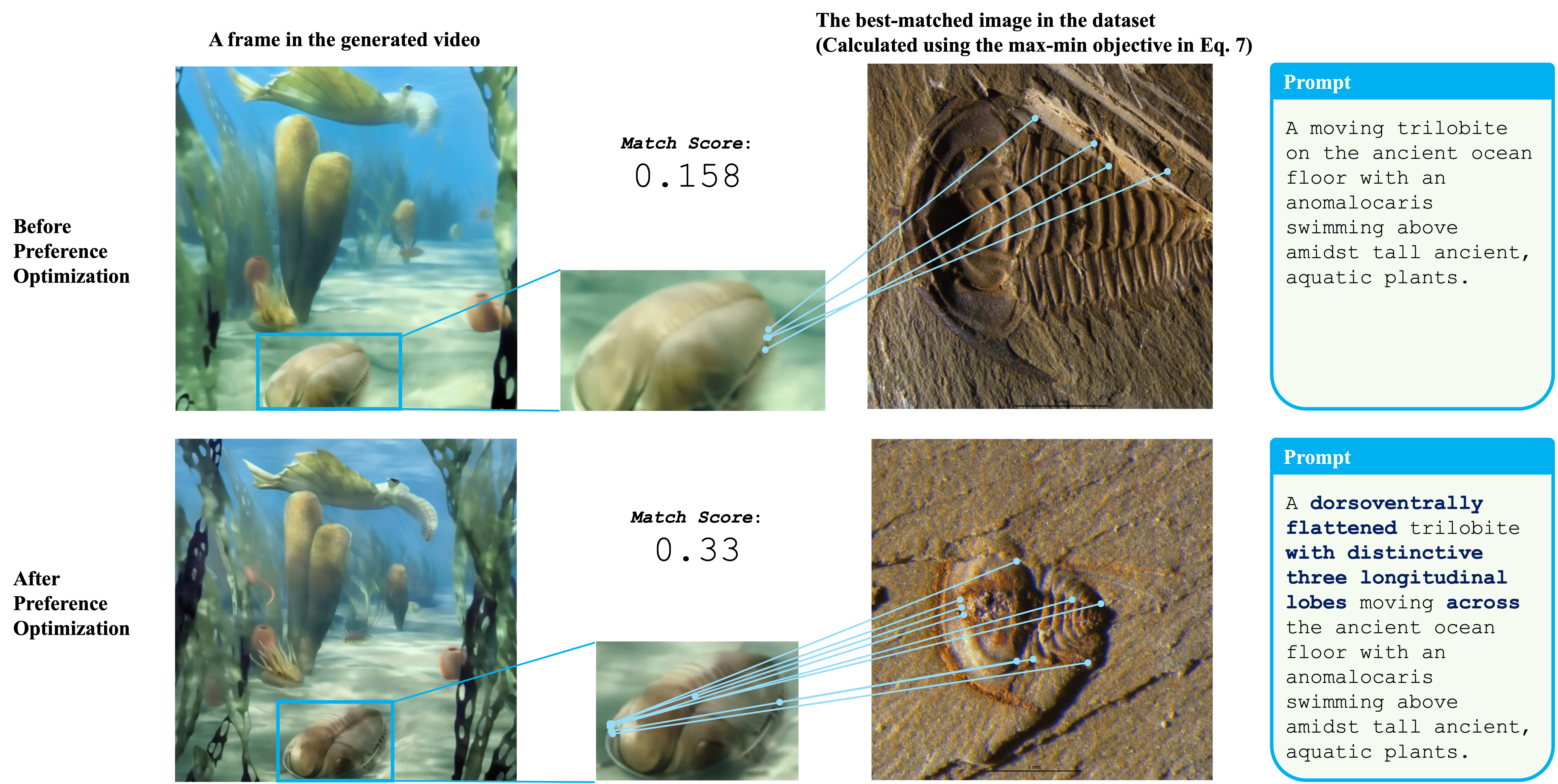}
    \caption[?]{The preference optimization of \wname~contributes to visual realism of the generated trilobites. What is \textcolor[rgb]{0,0.125,0.376}{\textbf{highlighted}} represents the changes in prompts after preference optimization. We can see that \textcolor[rgb]{0,0.125,0.376}{\textbf{\wname~learns to improve the quality of generated videos by adding more descriptions about the trilobite morphological details}}. The shown match score is the inverse of the distance from the most similar reference image ($1/\min_{r\in\mathcal{R}}D(x,r)$), which significantly improves after optimization. The first prompting image is courtesy of \cite{el2024rapid}. \emph{\textbf{Please note}: The reference images are used to enhance the visual details of the generated content. We show the reference image with the highest match score calculated by the ORB detector. The connecting lines represent matching points with similar local image features, and do \textbf{not} mean that the trilobite in the video matches the trilobite in the real fossil image.}}
    \label{fig:real_prompt_1}
\end{figure*}
\textbf{Visual Realism}.\ \ To ensure scientific rigorousness, we compare the visual details of the generated content against real samples of trilobite fossils from $\mathcal{R}$. For a video consisting of a set of frames, we expect that no frame has trilobites with morphological details deviating significantly from realistic data. To this end, we design a max-min objective:
\begin{equation}
r_a(y^0_n) = -\max_{x \in [f]} \min_{r \in \mathcal{R}}  D(x, r)~\label{equ:ra}
\end{equation}
Here, $D$ is a distance function that measures the morphological similarity between a generated trilobite and a reference image. We now explain the intuition of this similarity reward. The reference image set contains images of trilobite fossils from different growing stages, various sizes, different levels of preservation, and different geological periods. Therefore, we would have clear evidence that a generated trilobite is visually realistic if it is morphologically similar to at least one reference image. We capture this by the minimum operation in Eq.~\ref{equ:ra}. Then we try to find the frame that is the most different from the reference set, minimizing which could guarantee that no frame deviates too much from the reference set.

In practice, we use the ORB (Oriented FAST and Rotated BRIEF
(Binary Robust Independent Elementary Feature)~\cite{aglave2015implementation}) detector as the distance function $D$. The ORB detector is a fast and efficient feature detection algorithm. It combines the FAST keypoint detector and the BRIEF descriptor, providing robust performance suitable for extracting morphological details. We then use BF (Brute-Force)~\cite{antony2018implementation,noble2016comparison} method for matching the features. computes distances between every pair of descriptors, typically employing the Hamming distance for binary descriptors, as utilized in our case.


\subsection{Training the \wname}

Having defined the reward signals, we now introduce the third step, the training of \wname. We note that the rewards defined in the previous section are all negative, could be large in magnitude, and prone to noise, which indicates that these rewards may not be effective when used to train the \wname~LLM with algorithms like PPO~\cite{schulman2017proximal}, as they are sensitive to the specific values of the rewards. To tackle this problem, we propose ordering the prompts based on the rewards then applying preference optimization. Preference optimization has been proven its effectiveness in the broad literature of reinforcement learning from human feedback (RLHF)~\cite{ouyang2022training} and is more robust when reward values are noisy.

To be specific, we collect a training dataset $\mathcal{D}$ where each training sample contains a query $x$, a desirable generation $y_d$, and an undesirable generation $y_u$. Here, $r(y_d)>r(y_u)$. We use Kahneman-Tversky optimization (KTO)~\cite{ethayarajh2024ktomodelalignmentprospect} to train the \wname. Using $\lambda_y$ to denote $\lambda_D$ ($\lambda_U$) when $y$ is desirable (undesirable), where $\lambda_D$ and $\lambda_U$ are two constants, the KTO loss is:
\begin{equation}
\mathcal{L}_{\text{KTO}}(\theta^0) = \mathbb{E}_{x, y \sim \mathcal{D}} [\lambda_y - v(x, y)]
\end{equation}
where 
\begin{align}
r_{\theta^0}(x, y) &= \log \frac{\pi_{\theta^0}(y | x)}{\pi_{\text{ref}}(y | x)};\\
z_0 &= \mathbb{E}_{x' \sim \mathcal{D}} [KL(\pi_{\theta^0}(y' | x') \parallel \pi_{\text{ref}}(y' | x'))];\\
v(x, y) &= 
\begin{cases} 
\lambda_D \sigma(\beta (r_{\theta^0}(x, y) - z_0)) & \text{if } y \sim y_d | x; \\
\lambda_U \sigma(\beta (z_0 - r_{\theta^0}(x, y))) & \text{if } y \sim y_u | x.
\end{cases}
\end{align}
In the following section, we present examples showing how the KTO training improves the prompts.

After KTO training, \wname~parameters are updated to $\theta^1$, which leads to updated prompts $y^1$. $\theta^1$ can be further improved by running KTO given the new video generated by $y^1$. This process can be repeated until the new video is satisfactory enough. 



\section{Experiments}

In this section, we present experiments to test the effectiveness of our method. The experiments are designed to give both qualitative and quantitative evaluation of the generated videos. We compare against strong baselines, including previous work on text-to-animation (AnimateDiff~\cite{guo2023animatediff}) and powerful commercial text-to-video tools, Pika Labs~\cite{pika} and Gen-3~\cite{gen2}. We also carry out ablation studies to show the separate contribution of \wname~training and \vname~fine-tuning.


\subsection{Qualitative Results I: Visual Realism}

\textbf{Compare with baselines}.\ \  In Fig.~\ref{fig:real_compare_1}, we showcase a qualitative comparison of trilobite renderings produced by different models. The objective of the comparison is to evaluate each model's ability to generate realistic trilobites in various dynamic backgrounds.

\begin{figure*}
    \centering
    \includegraphics[width=\linewidth]{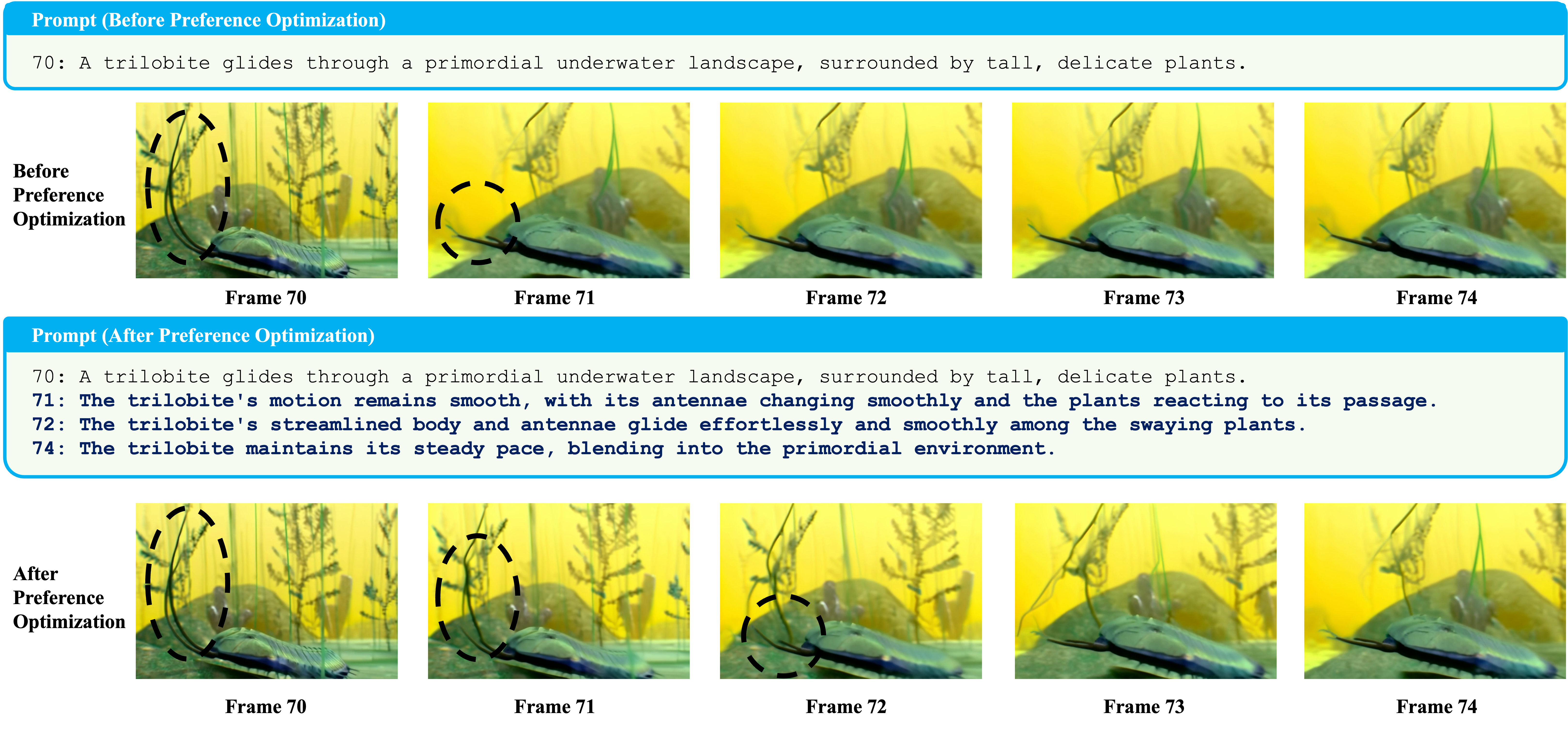}
\caption{A qualitative comparison before and after \wname~preference optimization, with a focus on the smoothness and continuity of the video. The \wname~learns to \textcolor[rgb]{0,0.125,0.376}{\textbf{add more prompts}} to impact the smoothness of the resulting video. The prompting image is courtesy of \cite{exampleImage1}.}
\label{fig:smooth_compare1}
\end{figure*}

The first row shows a scene featuring trilobites on the ocean floor with a volcanic eruption in the background. The Pika model generates a trilobite with unrealistic segmentation. The Runway model shows a more realistic structure but lacks in capturing the authentic texture of trilobite exoskeletons. The AnimateDiff model produces an oversimplified trilobite, and the main part of the image features a volcano. In contrast, the trilobites generated by our model display intricate segmentation, realistic texturing, and coloration that blends well with the naturalistic ocean floor setting, making them the most lifelike.

The depiction in the second row includes three trilobites among aquatic plants on the seabed. The Pika model's trilobites are not similar to any known types of trilobites. Runway's versions show better integration with the background but are still somewhat artificial in appearance. The trilobites by AnimateDiff lack depth and detail in texturing. Our model, however, shows trilobites with precise, well-defined segmentation and natural colors that harmonize with the underwater environment, enhancing the realism of the scene.

The scene of the third row captures a single trilobite moving along the ocean floor with a focus on the interaction with the environment, such as sediment displacement. Pika's rendition again lacks realism in appearance. AnimateDiff's trilobite appears round. Meanwhile, Our model produces a realistic trilobite interacting with its surroundings, showing sediment displacement that suggests a natural weight and presence in the water.

In summary, our model outperforms in creating trilobites with realistic anatomical features, textural fidelity, and appropriate environmental interactions compared to the other models. This qualitative analysis underscores the ability of our method in generating video content that closely mirrors the true appearance of trilobites.

\textbf{Compare with ablations}.\ \  Two components in our method contribute to the visual realism of the generated trilobites: we first fine-tune the T2A model and then carry out preference optimization for \wname. In Fig.~\ref{fig:real_prompt_1}, we show the influence of these two components. Specifically, we present two examples that demonstrate the results before and after preference optimization, with a focus on the visual quality of trilobite renderings in generated videos. Each example shows a frame from the video, accompanied by the most closely matching image from the dataset and the corresponding prompts.

In the first example (Fig.~\ref{fig:real_prompt_1}), before optimization, the video frame shows a trilobite on the ocean floor with a volcanic eruption in the background. The trilobite appears somewhat blended into the background, lacking distinct features, resulting in a low match score of 0.07. The prompt focuses on the general presence of trilobites amid a dynamic background. By contrast, after optimization, the optimized frame exhibits a trilobite with more emphasized and defined hard shells, enhancing its visibility and structural integrity against the complex background. This optimization is attributable to the updated prompt, which now specifically highlights the trilobite's hard shell. The match score significantly improves to 0.35, indicating a closer resemblance to the most similar reference image, which shows clearer and more detailed trilobite features.

In the second example (Fig.~\ref{fig:real_prompt_1}), before optimization, the original video frame captures a trilobite moving across the ocean floor with anomalocaris in the background. Initially, the trilobite lacks prominent distinguishing features, leading to a match score of 0.158. After preference optimization, the frame now shows the trilobite with enhanced distinguishing features, such as the longitudinal lobes and textural details, making it more realistic and akin to the reference image. Again, it is the updated prompt that leads to these changes, specifically by pointing out these features, contributing to a raised match score of 0.33.

In both cases, preference optimization led to adjustments in the model's rendering, focusing on enhancing specific features of the trilobites that contribute to greater visual realism. The targeted adjustments in the prompts post-optimization are pivotal in directing the model to produce outputs that not only adhere more closely to the reference images but also showcase more pronounced and authentic trilobite characteristics. This approach demonstrates the model's capability to adapt and refine its output by learning from preferences, ultimately yielding higher match scores and visually richer renderings.

\begin{figure*}
    \centering
    \includegraphics[width=\linewidth]{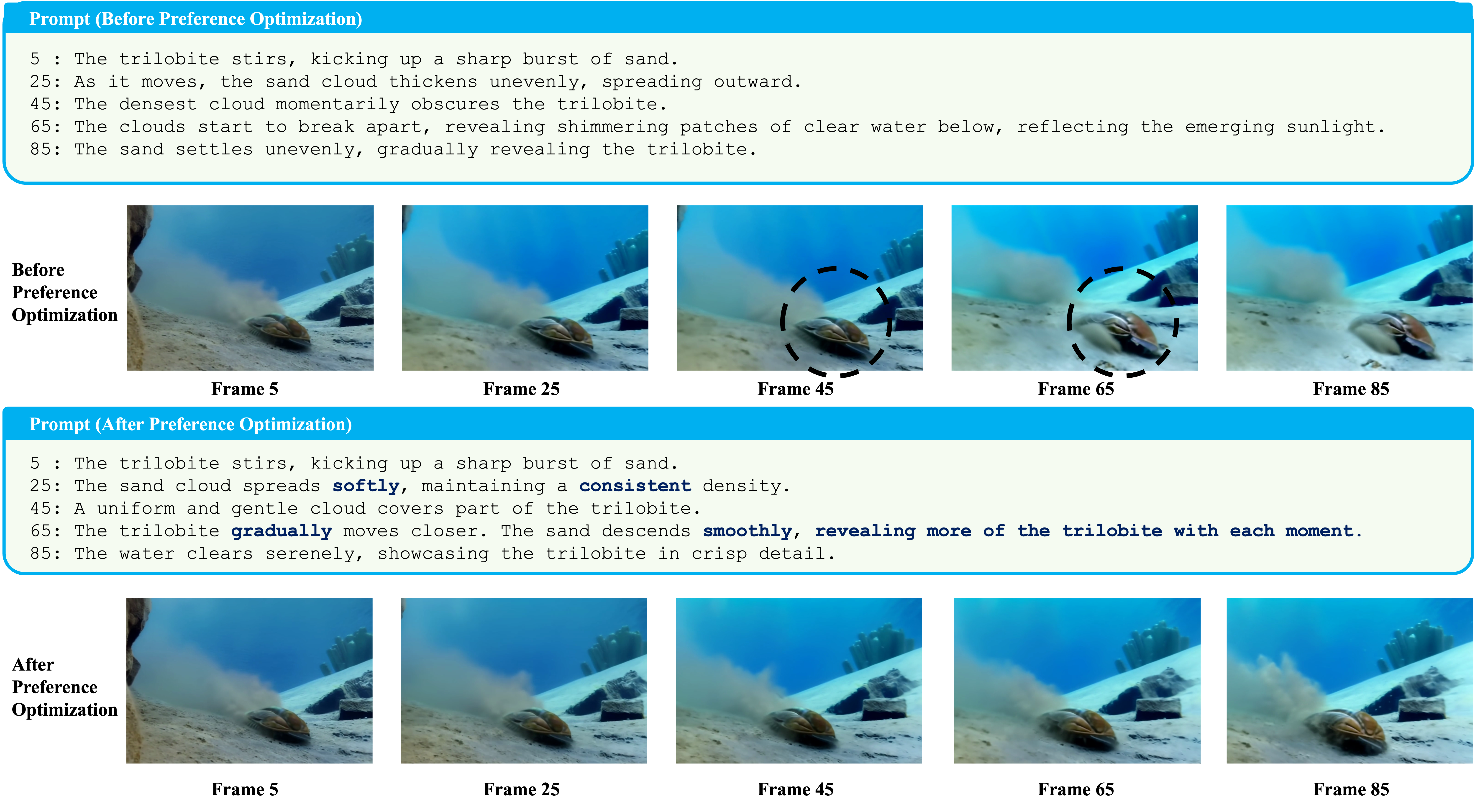}
\caption{Another qualitative comparison before and after \wname~preference optimization regarding video smoothness and continuity. The \wname~learns to \textcolor[rgb]{0,0.125,0.376}{\textbf{use words that indicate degree and process}} to enhance the resulting video smoothness.}
\label{fig:smooth_compare2}
\end{figure*}

\begin{figure*}[h]
    \centering
    \includegraphics[width=\linewidth]{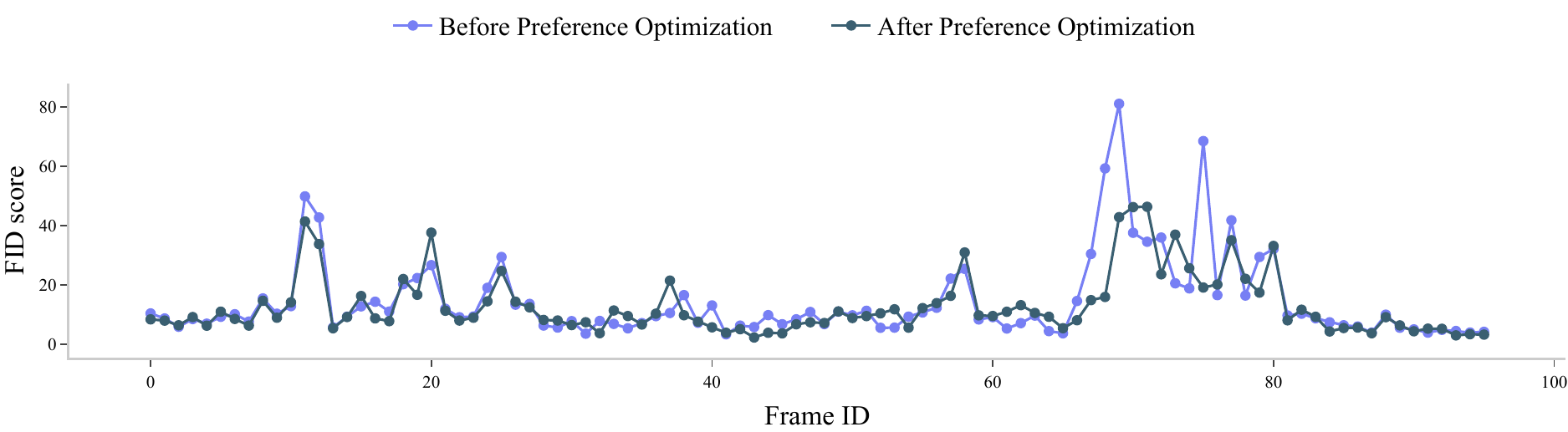}
\caption{Quantitative comparison: FID between adjacent frames. The x-axis represents the frame ID, ranging from 0 to 100, corresponding to the sequence of frames in the video. The y-axis quantifies the FID score, where a lower score indicates greater visual similarity and consistency between frames. The result shows that \wname~preference optimization effectively improves the smoothness of the generated video. }
\label{fig:smooth_curves}
\end{figure*}
\subsection{Qualitative Results II: Smoothness}

Fig.~\ref{fig:smooth_compare1} displays a series of frames. The sequences before and after preference optimization are shown. In the initial frames before optimization, the trilobite's movement appears somewhat jerky, especially its antennae. The corresponding prompt focuses on the trilobite's glide through the landscape. After preference optimization, the frames show a noticeable improvement in the fluidity of the trilobite's movement. The animation becomes smoother, with the trilobite seamlessly integrating into the motion of the surrounding plants. This creates a more naturalistic and visually appealing scene. The cause of this change is that the optimized prompt adds some frames and emphasizes the smooth, effortless glide of the trilobite and its streamlined body, highlighting how these characteristics should be reflected in the animation. This directive likely influenced the rendering process to focus on creating a smoother and more coherent movement pattern.

Fig.~\ref{fig:smooth_compare2} gives another example where the \wname~learns to add some words to enhance the video smoothness. The comparison clearly demonstrates that the changes in the prompts, post-optimization, lead to significant improvements in the smoothness of the video.


\subsection{Quantitative Results}

We conduct quantitative comparisons to further evaluate our method.

\textbf{Smoothness after KTO prompt training}.\ \ 
Fig.~\ref{fig:smooth_curves} illustrates the Fréchet Inception Distance (FID) scores between adjacent frames in a generated video sequence, comparing results before and after preference optimization. 

Before preference optimization, the blue line shows several peaks, particularly noticeable around frames 15, and 60-80, suggesting that the transitions between these frames are less smooth, with more noticeable visual discrepancies. After preference optimization, the dark line generally maintains lower FID scores throughout the sequence, with fewer and lower peaks compared to the blue line. This indicates that after optimization, the frames have greater visual consistency, and the transitions between them are smoother.

The overall trend in the graph demonstrates that preference optimization effectively reduces the FID scores across the majority of the video sequence. This improvement signifies that the video has become smoother post-optimization, with more consistent and visually coherent transitions between frames.

\textbf{User study}. \ \ We generate videos using four different methods and conduct a user study to evaluate their performance. Participants are asked to rate the videos on three criteria: smoothness, visual realism, and consistency with the prompt. We recruit participants for this study, each rating the videos on a scale from 1 to 4, where 4 indicates the highest possible score. This scoring system is equivalent to Average User Ranking (AUR), with higher scores indicating superior performance across the evaluated metrics.

Our method outperforms the other three methods in all evaluation criteria, indicating a significant improvement in video generation quality. This is evident from the higher scores across all three categories, confirming the effectiveness of our approach in producing videos that are smooth, visually realistic, and consistent with the given prompts. Particularly, the much higher scores regarding consistency with prompts achieved by our method highlight the effectiveness of our prompt learning method.

\begin{table}[]
    \centering
    \begin{tabular}{c|c|c|c}
    \midrule
    &Smoothness &Visual Realism &\makecell{Consistency\\ with prompt}\\
    \midrule 
         Ours &\textbf{3.41$\pm$ 0.21 }&\textbf{3.70$\pm$ 0.15}&\textbf{3.56$\pm$ 0.27}\\
         Gen3 &3.02$\pm$ 0.19&2.22$\pm$ 0.25&2.41$\pm$ 0.12\\ 
         Pika &2.47 $\pm$ 0.25 &2.67 $\pm$ 0.26 &2.58  $\pm$ 0.14 \\
         Animatediff &1.27$\pm$ 0.29&1.41$\pm$ 0.23&1.44$\pm$ 0.17\\
    \bottomrule
    \end{tabular}
    \caption{Quantitative comparison based on user study (mean $\pm$ var). A higher score indicates better performance.}
    \label{tab:user_study}
\end{table}

\section{Conclusion}

In conclusion, our study leverages advanced generative AI techniques to address the challenges of reconstructing trilobite behavior from fossil records. By integrating computational methods with paleontological research, we demonstrate the potential to enhance our understanding of these ancient creatures. Our proposed video generation framework, which incorporates realism and smoothness assessments into the workflow, produces more accurate and dynamic visualizations of trilobite movements. These enhanced animations not only improve scientific insights but also make the prehistoric world more accessible to the public. This interdisciplinary approach marks an advancement in both the fields of paleontology and (multi-modal) artificial intelligence, opening new avenues for future research and educational opportunities.

\section*{Acknowledgment}
We deeply appreciate Qiang Ou (China University of Geosciences, Beijing), Degan Shu (Northwest University, Xi’ an), Jian Han (Northwest University, Xi’ an), Meirong Cheng (Northwest University, Xi’ an) for generously providing  the image of trilobites that supported this study. 
\bibliographystyle{IEEEtran}
\bibliography{IEEEexample}

\vspace{12pt}

\end{document}